\documentclass[a4paper,conference]{IEEEtran}
\usepackage{cite}

\ifCLASSINFOpdf
  \usepackage[pdftex]{graphicx}
\else
\fi
\usepackage{amsmath}
\usepackage{amssymb}
\IEEEoverridecommandlockouts
\begin{document}
%
\title{Novel View Synthesis from only a 6-DoF Camera Pose by Two-stage Networks}

\author{\IEEEauthorblockN{Xiang Guo, Bo Li, Yuchao Dai $^{*}$, Tongxin Zhang, Hui Deng}
\IEEEauthorblockA{School of Electronics and Information,
Northwestern Polytechnical University, Xi'an, China}
\thanks{$^{*}$ Yuchao Dai is the corresponding author (daiyuchao@nwpu.edu.cn)}
}


%


\maketitle

\begin{abstract}
Novel view synthesis is a challenging problem in computer vision and robotics. Different from the existing works, which need the reference images or 3D models of the scene to generate images under novel views, we propose a novel paradigm to this problem. That is, we synthesize the novel view from only a 6-DoF camera pose directly. Although this setting is the most straightforward way, there are few works addressing it. While, our experiments demonstrate that, with a concise CNN, we could get a meaningful parametric model that could reconstruct the correct scenery images only from the 6-DoF pose. To this end, we propose a two-stage learning strategy, which consists of two consecutive CNNs: GenNet and RefineNet. GenNet generates a coarse image from a camera pose. RefineNet is a generative adversarial network that refines the coarse image. In this way, we decouple the geometric relationship between mapping and texture detail rendering. Extensive experiments conducted on the public datasets prove the effectiveness of our method. We believe this paradigm is of high research and application value and could be an important direction in novel view synthesis. 
\end{abstract}


%
\IEEEpeerreviewmaketitle

\section{Introduction}
Given a set of image-pose pairs, people could easily imagine a novel view of the object or scene. This is well known as the ``mental rotation'' which was found by psychologists more than 30 years ago. In the computer vision society, researchers imitate this character and propose a computational equivalency which is named as novel view synthesis (NVS). Given one or more input images of an object or a scene and the desired viewpoint transformation, the goal of NVS is to synthesize a new image capturing this novel view\cite{zhou2016view}. Existing work on NVS can be roughly classified into two categories: the geometric-based methods and the image generation-based methods. The geometric-based methods usually compute or estimate some geometric cues explicitly, e.g. the point cloud, depth image or appearance flow etc\cite{zhou2018stereo,Choi_2019_ICCV,reSfM,Meshry_2019_CVPR}. With this 3D information, the novel view synthesis problem is transferred as a projection or wrapping problem. The image generation-based methods generally train a parametric model to generate the target image directly. This kind of model needs the reference images and relative pose transformation as input in the training and inference time. Thus, these models essentially encode the reference image and relative pose transformation and decode the target image directly.

\begin{figure}[!t]
\centering
\includegraphics[width=3.0in]{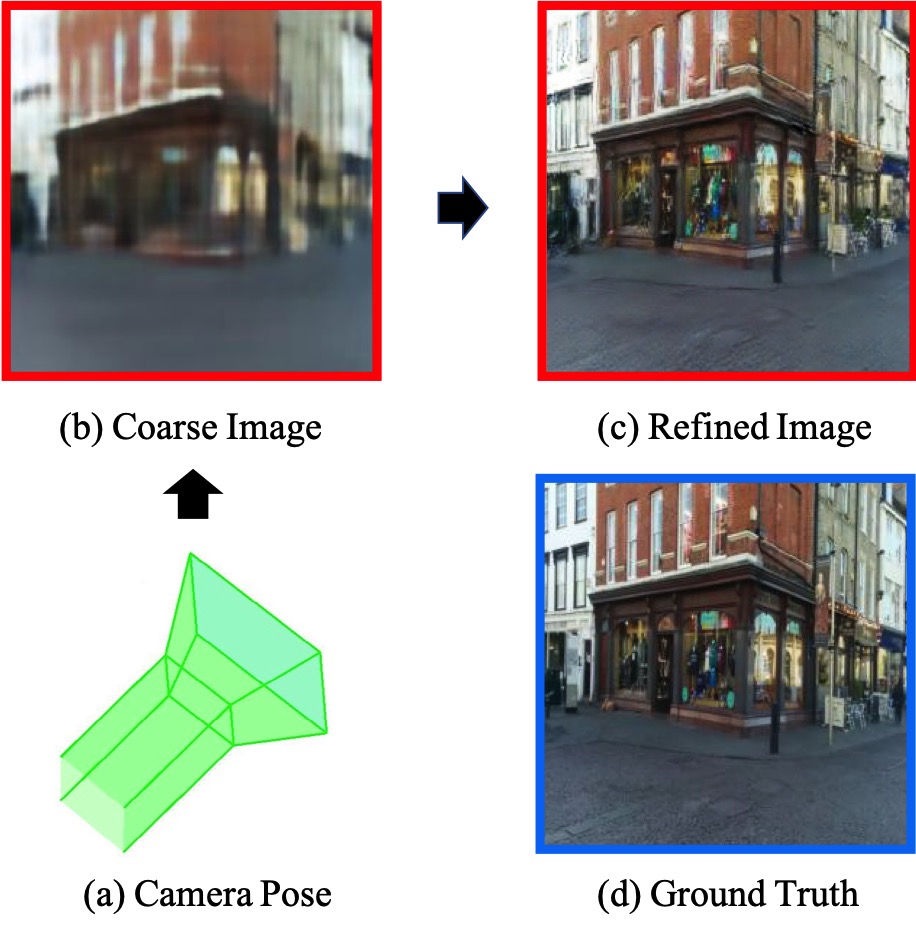}
\caption{Novel view synthesis from a 6-DoF pose. Our framework consists of two-stage networks. In the first stage, we could obtain a coarse image but with correct geometric structures. In the second stage, we refine the coarse image and get a high visual quality image.}
\label{show}
\end{figure}

Even though these approaches achieve success in generating novel view images, we would like to argue that if we could solve the novel view synthesis problem in a novel or more straightforward way, i.e. predict the novel view from a 6-DoF camera pose directly, as shown in Fig.~\ref{show}. It is worth noting that our work is partially inspired by the works of PoseNet\cite{PoseNet1}, which regress camera pose directly from a single image input \cite{MapNet,VidLoc,PoseNet1,Hourglass,AnchorNet}. To some extent, our work could be treated as the inverted progress of it. While, our task is much more difficult than it, considering the image contains rich information while the 6-DoF pose is rather abstract. Obviously, with this extreme setting, the models have to learn the correct geometric correspondence between the 6-DoF pose and images. More importantly, the models have to result in high-quality images with details and texture.

In addition, our work is also closely related to the work of \cite{tatarchenko2016multi}, which tries to generate the novel view of simulation objects from the attributes, e.g. colour, pose, or type, etc. But we would like to emphasize that our task is more difficult because we define the task on natural images. This makes our problem even harder. To the best of our knowledge, although the problem configuration is rather straightforward, this work is the first attempt to solve the novel view synthesis in this paradigm, especially for nature realistic images.

Besides the research interest (how well this can be done?), this novel problem configuration is relevant to many real-world applications and other related research topics such as virtual touring \cite{snavely2008modeling}, visual localization \cite{densevlad}, virtual/augmented reality, MVS \cite{MVS2}, depth estimation \cite{Li_2015_CVPR} and 3D reconstruction \cite{pami}. For example, it could help create full virtual reality environments based on historic images or video footage.

Considering the high abstract of the pose vector, it is difficult to generate the corresponding image directly. While we find a two-stage training strategy is rather effective to alleviate it. More specifically, our proposed framework consists of two consecutive networks: GenNet and RefineNet. GenNet is an up-sampling network that generates a coarse image according to the camera pose. RefineNet is a generative adversarial network based on U-Net that refines the coarse image into a fine detailed image. We find out that GenNet could learn the geometric correspondences from the pose to the image well. In addition, the RefineNet could add realistic texture and details to the output of GenNet. In this way, the final output is recovered with both correct geometric structure and high visual quality.

Our main contributions could be summarized as follow:

\begin{enumerate}

\item We present a novel problem configuration that is synthesizing novel views from a 6-DoF camera pose only. To the best of our knowledge, this is the first work trying to solve the novel view synthesis problem in such an extreme paradigm. Although the current performance is far from perfect, we believe it owns high research and application value and will be an important complement to the present framework of novel view synthesis.

\item  We propose a novel two-stage training strategy which is consisted of two consecutive networks: GenNet and RefineNet. GenNet is an up-sampling network that generates coarse images according to camera poses and RefineNet is a generative adversarial network based on U-Net that refines the coarse image into a fine detailed image. Surprisingly, our method could achieve comparable results with many other STOA methods that require the reference images or 3D model prior.

\item We conduct extensive experiments on public datasets. These experiments demonstrate the feasibility of this extreme setting and the effectiveness of our method. 

\end{enumerate}


\section{Related Work}
In this section, we would like to summarize some very related works, which are mainly on the topic of novel view synthesis, image-based localization, and adversarial learning.

\subsection{Novel View Synthesis}
The novel view synthesis problem, which generates new views of a scene from some known images, is useful to applications in virtual and augmented reality. In recent years, a lot of literature is devoted to solve this problem using deep learning techniques. In 2016, Zhou \emph{et al.}\cite{zhou2016view} propose the Appearance Flow method, which trains a convolutional neural network (CNN) to predict appearance flows that specify which pixels in the input view could be used to reconstruct the target view. And a fully-convolutional deep network is trained in \cite{zhou2018stereo} to infer the multi-plane image representation from stereo image pairs that can be used to synthesize novel views of the scene. Extreme view synthesis \cite{Choi_2019_ICCV} uses depth probability to model image depth which enables large extrapolation compared with \cite{zhou2018stereo}. Exploiting multi-view images, Sun \emph{et al.}\cite{sun2018multi} propose an end-to-end trainable framework to synthesize a novel view while \cite{liu2018geometry} focuses on addressing the problem of Single-Image Novel View Synthesis. 

On the other hand, \cite{reSfM, Meshry_2019_CVPR} are slightly different from the others. They use 3D point clouds information to get a realistic image at the target viewpoint without reference images. All these works need extra information (e.g. images or 3D point clouds) as input, while our method needs just the camera pose.

\subsection{Image based Localization}
In recent years, deep learning architectures are proposed to directly estimate image locations from the input images \cite{PoseNet1,PoseNetlstm}. Kendall \emph{et al.} originally propose Posenet, which utilizes GoogLeNet \cite{googlenet} and transfer learning, to enable direct image-to-pose regression and improve it by modeling uncertainty and introducing geometric constraints \cite{PoseNet3}. Other works introduce temporal information to smooth estimated results \cite{VidLoc} and geometric constraints to restrict network training \cite{MapNet}. Such direct deep regression networks achieve robust estimation in some difficult scenarios and real-time performance, although there is still a noticeable gap in terms of accuracy compared with traditional methods \cite{Limitation}. 

\subsection{Adversarial Learning}
The generative adversarial network (GAN) has achieved a great process in generating realistic images with high resolution and diversity. However, GAN model is known to be difficult to train with the model collapse problem and is unstable during training. While some works, like WGAN \cite{WGAN} and LSGAN \cite{LSGAN}, make modifications on loss function to achieve stabler training process, other works introduce gradient penalty \cite{WGAN-GP} and spectral normalization \cite{SNGAN} to stabilize the training process.

\begin{figure*}[!t]
\centering
\includegraphics[width=6.5in]{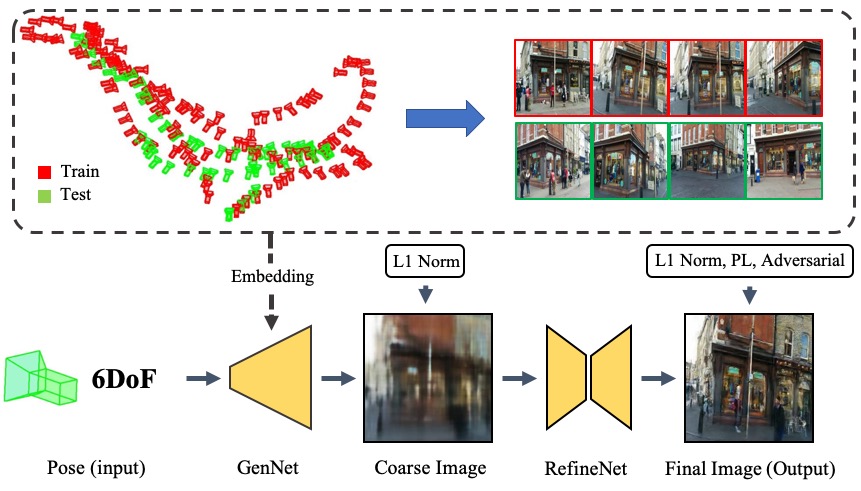}
\caption{Network Architecture: Our network contains two stages - GenNet and RefineNet. GenNet uses two Fully Connected layers to expand the input pose along the channel dimension. We use L1 norm for GenNet training and use L1 Norm, Perceptual loss (PL), and adversarial loss for RefineNet}
\label{fig_netarchi}
\end{figure*}


\begin{figure*}[!t]
\centering
\includegraphics[width=6.0in]{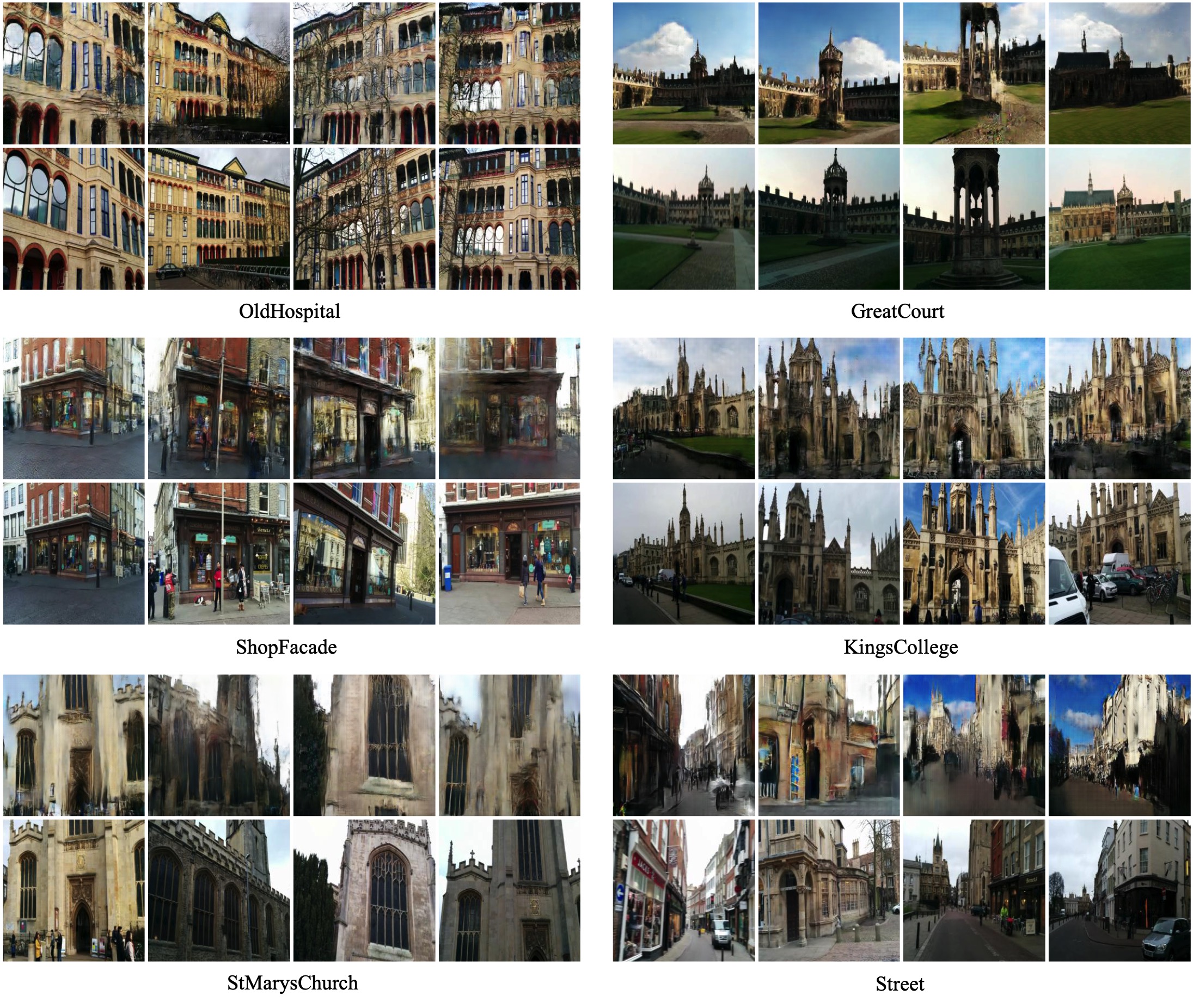}
\caption{Experimental results on the Cambridge Landmark dataset. For each scene, the first row contains the results of our proposed framework, and the second row contains corresponding ground truth images.}
\label{result1}
\end{figure*}

\section{Approach}
In this paper, we propose to deal with the novel problem of synthesizing novel view images from one 6-DoF camera pose only, which can be viewed as the invert of the PoseNet work \cite{PoseNet1} in visual localization. Our framework accepts a 6-DoF camera pose vector as the input, and generates a corresponding RGB image of the view directly. Specifically, the input camera pose consists of both camera rotation and camera translation. In this paper, we use the quaternion representation for rotation, which is a four-dimension unit norm vector, and the translation is represented as a three-dimension vector, thus the input camera pose vector $P$ is of 7 dimensions. Our novel task can be mathematically represented as $F_{\phi_G,\phi_R}(P): \mathbb{R}^{7} \to \mathbb{R}^{H\times W \times3}$, that is learning to synthesize an image from the input camera pose. 

We train two networks (GenNet and RefineNet) that connect in a sequence to synthesize an image from a camera pose. In the first stage, we use GenNet, $\phi_G$, to get a coarse image $I_c = \phi_G(P)$ from pose $P$. In the second stage, we use RefineNet $\phi_R$ to refine the coarse image $I_c$ to get a fine detailed image $I_r = \phi_R(I_c)$. 

The two fully connected layers in GenNet could encode the information of the scene which is the reason that our network could synthesize images without 3D model and reference images. The perceptual loss used in RefineNet could improve refinement performance.

\subsection{Network Architecture}
Our network architecture is shown in Fig.~\ref{fig_netarchi}, which consists of two consecutive sub-networks: GenNet and RefineNet. GenNet accepts a camera pose as input and generates the corresponding coarse image. Then RefineNet, a generative adversarial network (GAN) architecture, refines the coarse image to output the final fine detailed image. 

\noindent\textbf{GenNet.} The network architecture of GenNet consists of fully connected layers, up-sample layers, and convolutional layers. Inspired by \cite{Limitation}, the network expands the input along the channel dimension using two fully connected (FC) layer. Hence the dimension of input changes from $7\times1$ to $1024\times1$ after the final FC layer. Each FC layer is followed by a dropout layer with $p=0.2$. As described in \cite{Limitation}, the fully connected layer at the end of the absolute pose regression (APR) networks acts as an encoding function of the scene model. In this paper, as the work could be seen as an inverted process of APR, these two FC layers at top of the network could encode the information of the scene which enables the network generating images only depends on the input pose. There is also another way to expand the input by reshaping the output of the final FC layer into a $32\times 32$ feature map. According to our experiments, expanding input along the channel dimension leads to a better reconstruction of images. This may due to the wider structure introduced by the former method.

After two FC layers, the network attaches 8 up-sampling layers which are used to recover images from former scene embedding layers. Each up-sampling layers consist of a nearest neighbor up-sampling operation with a factor of 2, a convolution layer with 3 by 3 kernel, a batch norm layer, and the Relu activation. Finally, a convolution layer with Tanh activation is used to compress the number of channels into 3 to form a normal RGB image output. 

\noindent\textbf{RefineNet.} As shown in Fig.~\ref{fig_netarchi}, the RefineNet uses a U-net based GAN inspired by \cite{reSfM} revealing scenes by inverting structure from motion. We utilize the U-Net structure in pix2pix \cite{pix2pix}. \cite{pix2pix} shows low-level information from the input image is crucial for image transformation, and the skip connections between encoding layers and decoding layers in the generator could help this information flow through the network. As RefineNet aims to recover details of coarse images, these skip connects in the generator are obviously necessary for utilizing low-level features of the coarse image.

\subsection{Optimization}

\begin{figure*}[!t]
\centering
\includegraphics[width=6.0in]{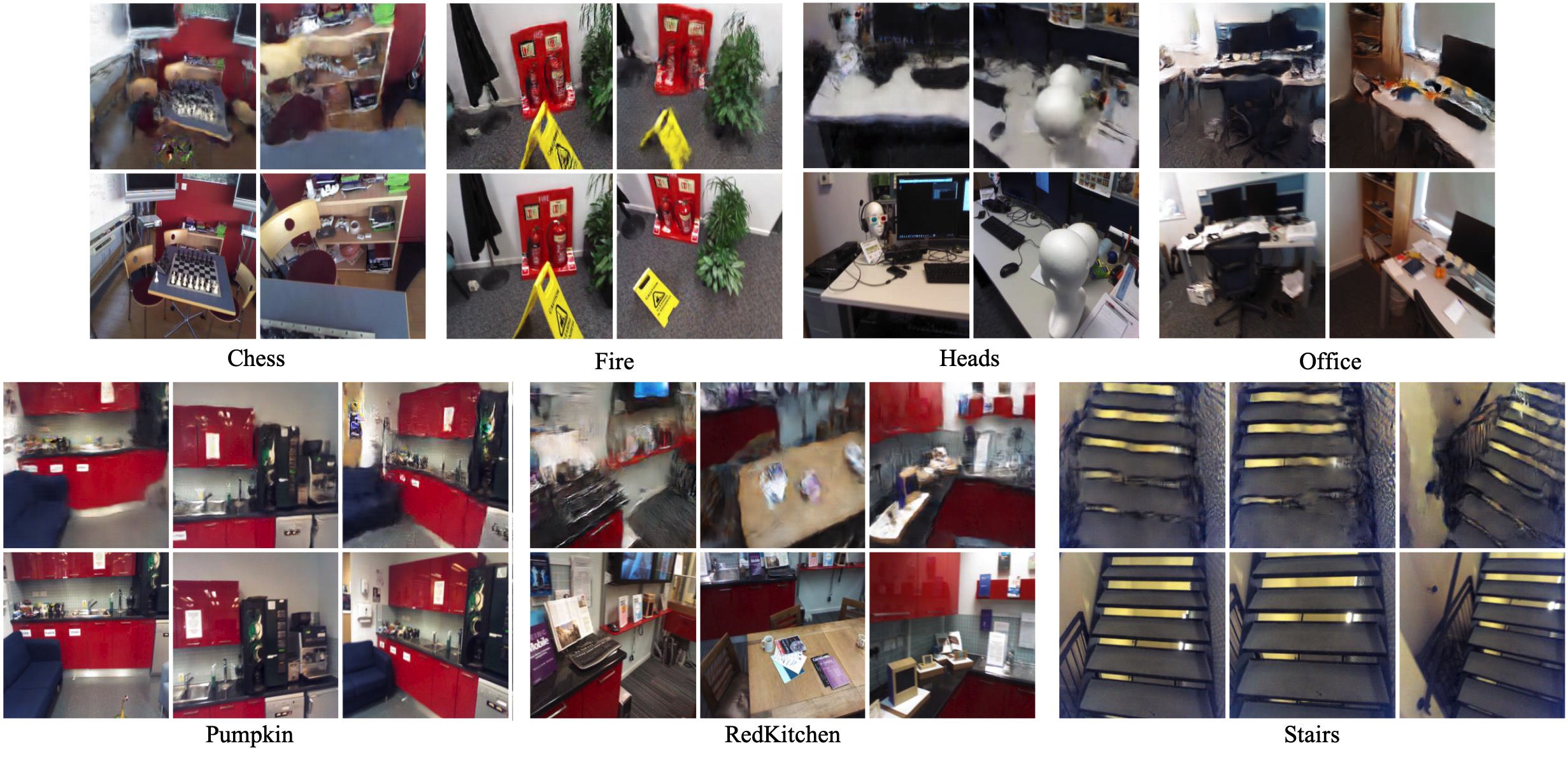}
\caption{Experimental results on the 7 Scenes dataset. For each scene, the first row contains the results of our method, and the second row contains corresponding ground truth images.}
\label{result2}
\end{figure*}

We train GenNet and RefineNet separately in our pipeline. For GenNet, given training pairs of input pose $P_{i} \in \mathbb{R}^{7}$ and corresponding target image $I_i \in \mathbb{R}^{256\times256\times3}$, we use the standard $L_1$ norm as GenNet loss function. 

RefineNet is trained based on the settings of \cite{pix2pix} and the parameters of GenNet are fixed during the training of RefineNet. The difference is that we do not use the dropout layer in order to produce deterministic results. Also, apart from the $L_1$ distance adding to the generator objective function in the original paper of \cite{pix2pix}, we additionally introduce the perceptual loss which is introduced in \cite{Perceptual}. According to \cite{Perceptual}, we define our style loss as follows: 
\begin{equation}
G_j^{\phi}(I)_{c,c^{'}}=\frac{1}{C_jH_jW_j}\sum_{h=1}^{H_j}\sum_{w=1}^{W_j}\phi_j(I)_{h,w,c}\phi_j(I)_{h,w,c^{'}},
\end{equation}
\begin{equation}
l_\textrm{{style}}(I_r,I_{gt})=\sum_{j=1}^{4}||G_j^{\phi}(I_r) - G_j^{\phi}(I_{gt})||_F^2,
\end{equation}
where $\phi_j(I)$ is the output of feature extractor network $\phi$ at the $j-$th layer of the input image $I$, and the feature map size is $C_j\times H_j\times W_j$. For output image of the GAN's generator $I_r$ and ground truth image $I_{gt}$, the style loss (2) is the sum of squared Frobenius norm of the difference between the Gram matrices calculated by (1). In our case, we use Vgg16 \cite{vgg} as our feature extractor network, and for $j=1,2,3,4$, the layers are relu1\_2, relu2\_2, relu3\_3, relu4\_3.

Furthermore, we define the content loss as follows:
\begin{equation}
l_\textrm{{content}}(I_r,I_{gt})=\frac{1}{C_2H_2W_2}||\phi_2(I_r) - \phi_2(I_{gt})||_2^2.
\end{equation}

We use the output of the second layer relu2\_2 to get content loss. As described in \cite{Perceptual}, content loss encourages the network to encode perceptual information of the dataset which enables our network to refine the coarse images with visual similar details.

Finally, we get our objective for GAN
\begin{equation}
\begin{split}
G^*=\arg&\mathop{\min}_{G}\mathop{\max}_{D}~l_{\textrm{cGAN}}(G,D)~ + \\&\lambda_1l_\textrm{{L1}}(G)+\lambda_2l_{\textrm{style}}(G) + \lambda_3l_\textrm{{content}}(G),
\end{split}
\end{equation}
where $l_{\textrm{cGAN}}(G,D)$ is a standard conditional GAN objective, $l_\textrm{{L1}}$ is standard $L_1$ norm.

\section{Experiments}

\begin{table*}[!t]
\renewcommand{\arraystretch}{1.3}
\centering
\scriptsize
\begin{tabular}{|c|c|c|c|c|c|c|c|c|c|}
\hline                    
~~& \multicolumn{3}{|c|}{GreatCourt} & \multicolumn{3}{|c|}{KingsCollege} & \multicolumn{3}{|c|}{OldHospital}\\
\hline
~~ & Coarse & Refined (PL) & Refined (w/o PL) & Coarse & Refined (PL) & Refined (w/o PL) & Coarse & Refined (PL) & Refined (w/o PL)\\
\hline
SSIM & 0.4361 &	0.3916 & 0.3903 & 0.2953 & 0.2397 & 0.2370 & 0.1897 & 0.1300 & 0.1308\\ 
\hline
PSNR & 11.5266 & 11.3886 & 11.3648 & 12.9879 & 12.5296 & 12.5288 & 12.4578 & 11.5834 & 11.5353\\
\hline
L1 & 0.2023 &	0.2050 &	0.2055 &	0.1706 &	0.1803 &	0.1800 &	0.1785 	&0.1977 &	0.1987 \\
\hline
Brenner &	191.3672 &	466.3329 &	471.1897 &	232.5756 &	744.3240 &	752.2070 &	180.5834 &	1360.8733 &	1368.9296 \\
\hline
\hline
~~& \multicolumn{3}{|c|}{ShopFacade} & \multicolumn{3}{|c|}{StMarysChurch} & \multicolumn{3}{|c|}{Street}\\
\hline
~~ & Coarse & Refined (PL) & Refined (w/o PL) & Coarse & Refined (PL) & Refined (w/o PL) & Coarse & Refined (PL) & Refined (w/o PL)\\
\hline
SSIM &	0.2495 &	0.1768 & 	0.1372 &	0.2962 &	0.2172 & 	0.2158 &	0.2255 &	0.1303 &	0.1372 \\
\hline
PSNR &	12.7433 &	11.9713 &	11.9403 &	12.8790 &	12.4334 &	12.3585 &	10.3989 &	9.7498 & 	9.6893 \\
\hline
$L_1$ &	0.1818 	& 0.1951 &	0.1955 &	0.1777 &	0.1855 &	0.1867 &	0.2456 &	0.2633 &	0.2653 \\
\hline
Brenner	& 113.3604 &	692.7051 &	697.9377 &	117.7310 &	538.1169 &	619.7828 	&109.2868 &	870.7560 &	997.1627 \\
\hline
\end{tabular}
\caption{Quantitative evaluation of the synthesized images quality. Three measure methods with reference image: SSIM, PSNR, $L_1$ norm and one method without reference image: Brenner. Coarse means the coarse images generated by GenNet and Refined means refined image by RefineNet with or without Perceptual Loss (PL).}
\label{imagequa}
\end{table*}

In this section, we report extensive experiments evaluation of our method. We implement our pipeline with pytorch and use Adam optimizer to train both our GenNet and Refinet with $\beta_{1}=0.5$ and $\beta_{2}=0.999$. We use the same training and testing split in the dataset we used, and the testing sets are never involved in neither the training of GenNet nor RefineNet. Also, images in train sets and test sets in datasets are sampled from different trajectories.

\subsection{Dataset}
We use the Cambridge Landmarks \cite{PoseNet1} and 7-Scenes \cite{7Scenes}. Cambridge Landmarks is a large urban relocalization dataset containing over 12,000 images with 6 scenes around Cambridge University. 7-Scenes is an indoor dataset with 7 scenes containing around 7000 tracked RGB-D frames. Cambridge Landmarks and 7-Scenes cover indoor and outdoor scenarios with variances scales of scenes and different content. Both datasets contain several sequences capturing the same scene with different moving patterns. We separate the sequences into the test and training set, so that test and train set share no image in one sequence and have different poses in general, as shown in Fig.~\ref{fig_netarchi}.

\begin{figure}[!t]
\centering
\includegraphics[width=3.2in]{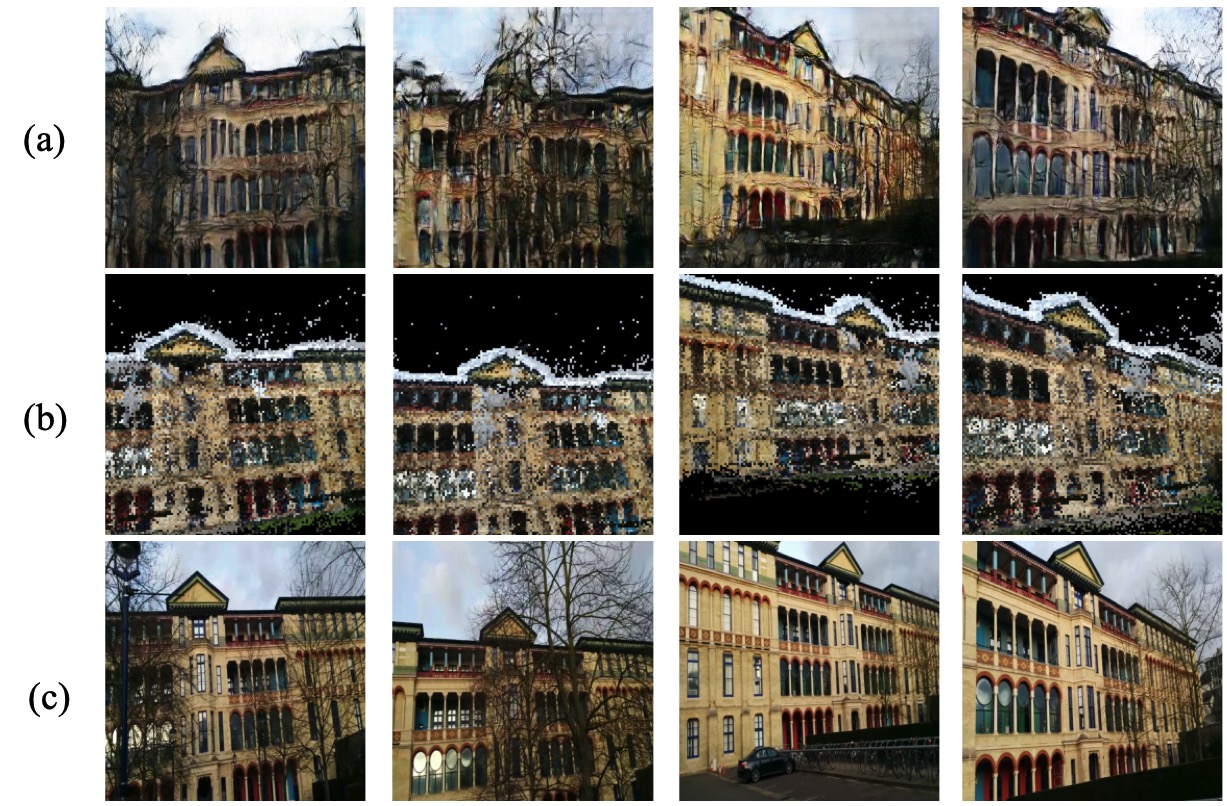}
\caption{Comparison with 3D model projection based image synthesize. (a) The results of our method, (b) The projections of the 3D point-cloud model, (c) Ground truth images.}
\label{projection}
\end{figure}

\subsection{Results}
We show qualitative results of our methods on the Cambridge Landmarks in Fig. \ref{result1} and 7-Scenes in Fig.~\ref{result2}. As we can see from Fig.~\ref{result1} and Fig.~\ref{result2}, both datasets involve large view change and object variance, while the Cambridge Landmarks even have weather and light changes. This leads to a significant weather differences in the sky and illumination differences of the buildings between synthesized images and ground truth images in GreatCourt and KingsCollege as shown in Fig.~\ref{result1}. The Street in Cambridge Landmarks is the most difficult scene with the most changes of the views and the longest trajectory, and it could be seen in Fig.~\ref{result1} that the network has more difficulties in synthesizing images in the Street scene.

Generally, 7-scenes has more details and objects with irregular shapes, and results in a general lower synthetic image quality. However, image detail is not the only factor that determines the difficulty. For example, the Stairs scene in Fig.~\ref{result2} have relatively simple and regular objects, but the network has difficulties in learning such textures of the stairs.

We also show quantitative evaluations of the results of our methods on the Cambridge Landmarks in Table \ref{imagequa}. We use three methods requiring reference image (ground truth image) which are SSIM, PSNR, and L1 norm, as well as Brenner which does not require reference image to measure image quality. The evaluation metric is defined as follow:
\begin{equation}
SSIM(I,I_{gt})= \frac{(2\mu_{I}\mu_{I_{gt}}+c_1)(2\sigma_{I,I_{gt}}+c_2)} {(\mu_{I}^2 + \mu_{I_{gt}}^2+c_1)(\sigma_{I}^2 + \sigma_{I_{gt}}^2+c_2)},
\end{equation}
where $I$ could be either coarse image $I_c$ or refined image $I_r$, $\mu$ is the mean of the image, $\sigma_{I,I_{gt}}$ is the covariance of image $I$ and ground truth image $I_{gt}$, $\sigma^2$ is the variance of the image, $c_1,~c_2$ are constants.
\begin{equation}
PSNR(I, I_{gt}) = 20\log_{10}\left(\frac{MAX_I}{\sqrt{MSE(I,I_{gt})}}\right),
\end{equation}
where $MSE(I,I_{gt})$ is the mean square error between image $I$ and ground truth $I_{gt}$, and $MAX_I=1$ as we scale image to [0, 1] to calculate PSNR.
\begin{equation}
Brenner(I) = \sum_{h=0}^{253}\sum_{w=0}^{255}|f(h+2,w)-f(h,w)|^2,
\end{equation}
where $f(h,w)$ is the gray scale value at pixel $I(h,w)$.

\begin{figure}[!t]
\centering
\includegraphics[width=3.2in]{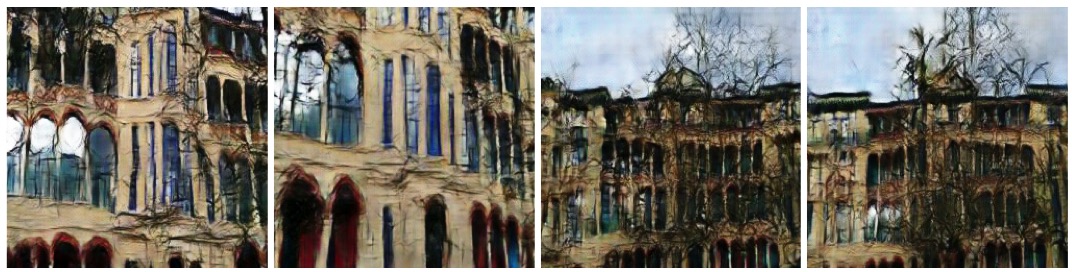}
\caption{Synthesize image from novel view points which do not exist in train and test trajectories.}
\label{newview}
\end{figure}

\begin{figure}[!t]
\centering
\includegraphics[width=3.2in]{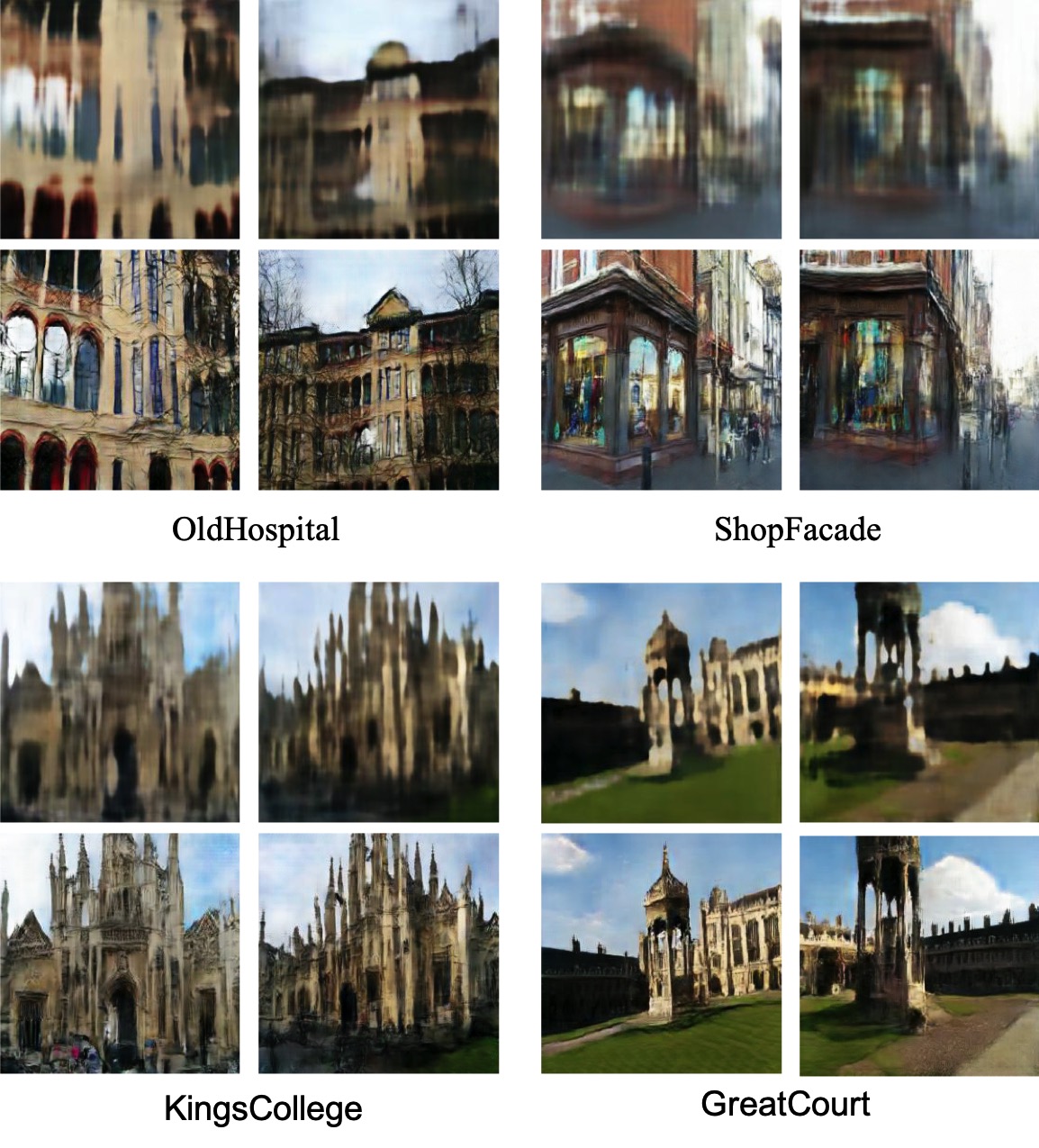}
\caption{Effect of RefineNet. For each scene, the first row is coarse images produced by GenNet and the second row is refinement results of RefineNet.}
\label{proveRefineNet}
\end{figure}

\begin{figure*}[!t]
\centering
\includegraphics[width=6.0in]{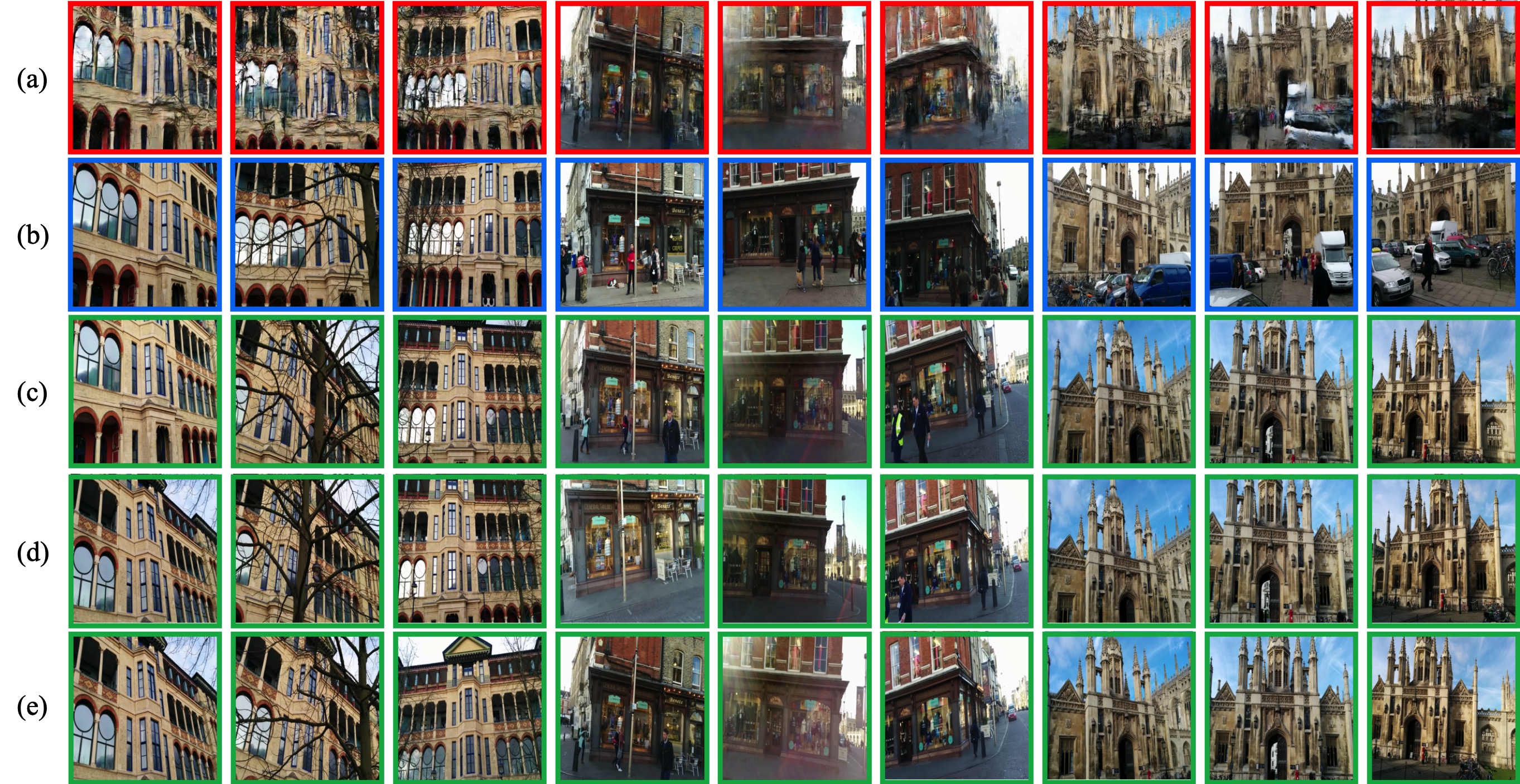}
\caption{The pose accuracy of generated images in OldHospital, ShopFacade, and KingsCollege. (a) synthetic images, (b) ground truth images, (c)(d)(e) top3 closest image to the corresponding pose in the training dataset in descend order.}
\label{accuracy}
\end{figure*}

\noindent\textbf{Comparison with 3D point clouds based view synthesis} In order to compare with view synthesis with 3D models, we set up a sample baseline. We use the Colmap \cite{colmap} to generate a dense reconstruction of OldHospital and re-project the dense 3D points according to camera poses. According to Fig. \ref{projection}, we could see that the re-projections are filled with areas that have no data and could not synthesize points that are not detected as interesting points during the reconstruction process. These experiments demonstrate the superiority of our proposed framework in capturing the global scene structure and achieving high-quality image synthesis with fine details.

\noindent\textbf{Novel View Synthesis} We generate some synthesized images from novel views that are not in train and test trajectories and show them in Fig. \ref{newview}.

\section{Ablation Studies}
In this section, we establish ablation studies to explore the strength of RefineNet and perceptual loss used in GAN training. Meanwhile, we discuss issues including pose accuracy of generated image and show qualitative results of novel view synthesis.

\subsection{RefineNet}
RefineNet is crucial for our pipeline in order to get fine detailed synthetic images. As shown in Table \ref{imagequa}, for Brenner method which is a traditional measurement of image quality without reference images, the results of images are increased significantly after the RefineNet in all scenes. Even there is a noticeable drop in the results of the other three measure methods that require reference image, this does not mean the image quality decrease after RefineNet and the reason will be explained in the following sections.

Fig. \ref{proveRefineNet} shows the comparison of intermediate coarse images and final images refined by RefineNet which present great improvement in visual quality. As shown in Fig. \ref{proveRefineNet}, RefineNet could refine details and adjust shapes of objects in coarse images. For example, RefineNet could turn shade in OldHospital into trees and enhance the glass effect of the windows significantly in both OldHosital and ShopFacade. In KingsCollege and GreatCourt, most work of RefineNet is to clean and straighten out of the edges of those long, narrow objects like spires and columns. This effect is also obvious in ShopFacade.

\subsection{Perceptual Loss}
\begin{figure}[!t]
\centering
\includegraphics[width=3.2in]{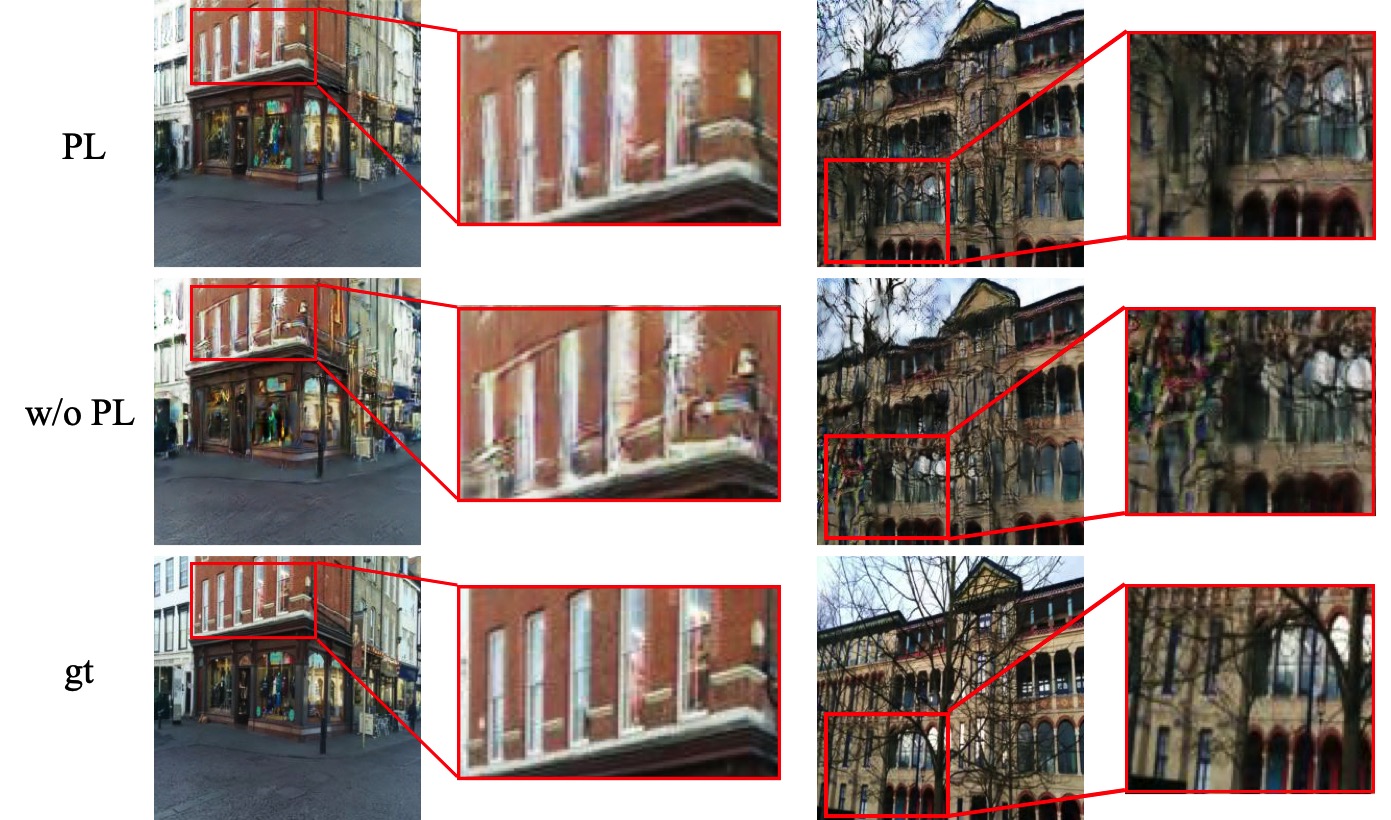}
\caption{Effect of Perceptual Loss. The first row is the results of RefineNet trained with perceptual loss. The second row is the results of RefineNet trained with the same setting but without perceptual loss. The bottom is ground truth images}
\label{provePL}
\end{figure}

The perceptual loss we used in RefineGan training makes a noticeable improvement of image quality compared with RefineGan without perceptual loss. In Table \ref{imagequa}, we can see a general improvement in terms of SSIM, PSNR, and L1 in most scenes. For Blenner, the score decreased once applied perceptual loss, as perceptual loss eliminates the high-frequency noise in images. We show some details in Fig. \ref{provePL}. As we can see, with perceptual loss, there are fewer artifacts and the edges of the building are cleaner and clearer. 

\subsection{Pose Accuracy of Generated Images}
The purpose of this paper is to generate a corresponding image according to the pose in an end-to-end manner. Hence, it is necessary to find a way to measure the accuracy of the synthetic image in terms of visual localization. A natural way is to locate the synthetic image using visual localization methods to get the camera pose and measure the difference between this pose and the ground truth pose. However, synthetic images are always filled with noise, distortion, and unrealistic artifacts, which could make it very difficult to generate a reasonable pose out of them just like humans do. Here, we provide some visualization in terms of pose accuracy of synthetic images. 

In Fig. \ref{accuracy}, the first row is synthetic images, and corresponding ground truth images are in the second row. The following rows contain the top3 nearest images to the test pose in training sets. As we can see, the network tends to generate images according to the view which it has seen during training which is particularly obvious in ShopFacade showing in the middle of the figure. This is more likely to be a retrieval behavior, which means the network tends to retrieve a similar pose in the training set and produce the corresponding images rather than exploring the geometric constraints between the scene pixel and the pose. This observation is similar to the conclusion of \cite{Limitation}. Even though, we could also obverse that the network does try to modify images in order to narrow down the difference between poses of the synthetic image and the ground truth. For example, in the first column, the network moves the image towards right and add an additional window based on the knowledge of training images. This network character could be seen in all of these examples, even in ShopFacade. 

In Fig. \ref{accuracy}, we can notice there is a significant gap between the pose of synthetic image and ground truth pose which is caused by the network's limited generalization ability. As test images are never used during the train of both GenNet and RefineNet, RefineNet is not trained in terms of close this pose gap. Due to the existence of the pose gap, the refinement of the coarse image enhances the differences between refined images and ground truth images in terms of SSIM, PSNR and $L_1$ in Table \ref{imagequa} which cause a performance drop after refinement.

\section{Conclusions}
In this paper, we have proposed a new problem configuration in novel view synthesis area, synthesizing images of a scene with one given 6-DoF camera pose only, without reference images, 3D model or other information. Our proposed framework consists of two networks that are used to generate a coarse image and then refine it into fine detailed image. Extensive experiments show promising results of our proposed framework in generating visually pleasant images. Furthermore, as a reverse process of Posenet \cite{PoseNet1}, our proposed network shows an ability of reasoning the relations between image and pose, while has a retrieval behavior. This could be an interesting future direction in investigating the network behavior for end-to-end visual localization.

\noindent\textbf{Acknowledgment.}
This work was supported in part by Natural Science Foundation of China (61871325, 61671387) and National Key Research and Development Program of China under Grant 2018AAA0102803.

{
\bibliographystyle{IEEEtran}
\bibliography{InvertPoseNet_Ref}
}
%



\end{document}


\title{InvertPoseNet: Learn to Synthesize Image from Arbitrary Viewpoint \\ -- Supplementary Material --}

\author{First Author\\
Institution1\\
Institution1 address\\
{\tt\small firstauthor@i1.org}
\and
Second Author\\
Institution2\\
First line of institution2 address\\
{\tt\small secondauthor@i2.org}
}

\maketitle
\section{Implementation Details}
In this supplementary material, we give a more detailed description of our network and training procedure.

\subsection{Architecture}
Our network architecture includes two sub-networks: GenNet and RefineNet. The input of the network is a 7 dimensional vector, which former 3 dimensions represent the spacial coordinates of the camera pose and later 4 dimensions is a quaternion representing the rotation of the camera. 

The architecture of GenNet is $F2048-F1024-UC512-UC256-UC256-UC256-UC256-UC256-UC128-UC64-C3$. $FN$ means fully connected layer with output dimension $N$ without bias, followed by a batch norm, ReLU activation and a dropout layer with p = 0.2. $UCN$ means a nearest neighbor upsampling with the factor equal to 2 and a convolutional layer with $N$ kernels of 3 $\times$ 3, stride and padding equal to 1, followed by batch norm and ReLU activation. $C3$ means a convolutional layer with 3 kernels of 1$\times$ 1, followed by Tanh activation. Suppose batch size is $bs$, the ouput of $F1024$ is resized from $bs\times1024$ to $bs\times1024\times1\times1$ along channel dimension before fed into the next layer.

We use the pytorch version of pix2pix as our RefineNet from \url{https://github.com/junyanz/pytorch-CycleGAN-and-pix2pix}. We set the model to pix2pix, load size to 256, and not using dropout layer in order to generate deterministic results, which is very different from the general purpose of GAN networks. 

\subsection{Optimization}

In order to train GenNet work, we use Adam optimizer with $\beta_{1}=0.5$ and $\beta_{2}=0.999$ on a NVIDIA GeForce GTX 1080Ti card. We set batch size to 48 and learning rate to $1e-4$. The training converge after around 1000 epochs for Cambridge Landmark scenes and 500 epochs for 7-Scene. During the training of GreatCourt, we remove all batch norm layers in GenNet, as it cause significant drop of performance of the network in evaluation mode. This may cause by the different distribution of test and training data, and the batch norm layer cannot model the distribution of the test data correctly learning from training data. 

For the training of RefineNet, we use the default settings except the perceptual loss we introduced to the network objective. We set $\lambda_2 = 5\times1e4$ for $l_{\textrm{style}}$ and $\lambda_3=10$ for $l_\textrm{{content}}$.

\section{Additional Results}
We show more qualitative results on both Cambridge Landmark and 7-Scene in Fig. \ref{supply}. Also, we show some failure cases. As shown in Fig. \ref{fail}, there are generally two kinds of failure cases for our networks. For the first type, our network may fail to generate meaningful scene, like the second to fourth columns in Fig. \ref{fail} or fail to synthesize some parts of objects, like the first column in Fig. \ref{fail}. In the second case, the network could generate images that the camera poses of these synthetic images are different with ground truth significantly.

\textbf{Supplementary Video} Finally, we provide two synthetic videos on OldHospital and ShopFacade in Cambridge Landmark, and both videos consist of 105 images. We generate the video frame by frame base on a novel trajectory and do not produce any local or global optimization among these frames. For OldHospital, we generate a trajectory that go straight along the hospital with a constant quaternion. For ShopFacade, as the shop locate at a corner of a street, we generate a curve trajectory going from one side of the shop to another side. For quaternion, we change quaternion every 3 frames, in order to keep camera watching the shop.

\begin{figure*}[!t]
\centering
\includegraphics[width=6.5in]{supply.png}
\caption{More Qualitative results on Cambridge Landmark and 7-Scene. (a): coarse images generated by GenNet, (b): final images generated by our network (refined by RefineNet), (c): corresponding ground truth images}
\label{supply}
\end{figure*}

\begin{figure*}[!t]
\centering
\includegraphics[width=6.5in]{fail.png}
\caption{Failure Case. Top row: results of our network, Bottom row: corresponding ground truth. Failure case can be generally divided into two types: failure to synthetic meaningful scene or parts of objects (in {\color{red} red}), failure to synthetic scene with reasonable camera pose close to ground truth (in {\color{green} greed})}
\label{fail}
\end{figure*}